\documentclass[10pt,twocolumn,letterpaper]{article}

\usepackage{iccv}
\usepackage{times}
\usepackage{epsfig}
\usepackage{graphicx}
\usepackage{amsmath}
\usepackage{amssymb}
\usepackage{booktabs} 
\usepackage{diagbox}
\usepackage{caption}
\usepackage{graphicx}
\usepackage[ruled,linesnumbered]{algorithm2e} 
\usepackage{subcaption}

\usepackage[breaklinks=true,bookmarks=false]{hyperref}

\iccvfinalcopy 

\def\iccvPaperID{2891} 

\begin{document}

\title{Stochastic Region Pooling: Make Attention More Expressive}

\author{Mingnan Luo, Guihua Wen\thanks{corresponding author}, Yang Hu\thanks{equal contribution with Guihua Wen}, Dan Dai, Yingxue Xu\\
{School of Computer Science \& Engineering, South China University of Technology}\\
{Panyu, Guangzhou, Guangdong, China}\\
{\tt\small\{csluomingnan@mail.,crghwen@,cssuperhy@mail.,csdaidan@mail.,201530381885@mail.\}scut.edu.cn}
}
%

\maketitle

\begin{abstract}
Global Average Pooling (GAP) is used by default on the channel-wise attention mechanism to extract channel descriptors. However, the simple global aggregation method of GAP is easy to make the channel descriptors have homogeneity, which weakens the detail distinction between feature maps, thus affecting the performance of the attention mechanism. In this work, we propose a novel method for channel-wise attention network, called Stochastic Region Pooling (SRP), which makes the channel descriptors more representative and diversity by encouraging the feature map to have more or wider important feature responses. Also, SRP is the general method for the attention mechanisms without any additional parameters or computation. It can be widely applied to attention networks without modifying the network structure. Experimental results on image recognition datasets including CIAFR-10/100, ImageNet and three Fine-grained datasets (CUB-200-2011, Stanford Cars and Stanford Dogs) show that SRP brings the significant improvements of the performance over efficient CNNs and achieves the state-of-the-art results.
\end{abstract}

\section{Introduction}


Convolutional neural network (CNN) is an effective method to solve the computer vision tasks~\cite{krizhevsky2012imagenet,simonyan2014very,he2016deep}. Furthermore, combining it with attention mechanisms can better solve them~\cite{chen2017sca,wang2017residual,yang2016stacked,nguyen2018attentive}, such as channel-wise attention networks~\cite{hu2018squeeze,li2018harmonious,hu2018gather,woo2018cbam,zhu2019stacked}. They usually use Global Average Pooling (GAP) to squeeze the entire feature map into a descriptor~\cite{hu2018squeeze,hu2018competitive,li2018harmonious,woo2018cbam}. However, GAP tends to ignore the detail area in feature map with lower magnitude, which easily leads to the homogeneity of the channel descriptors. To alleviate this problem, some researchers combine channel-wise attention with spatial attention to make the attention module to pay attention to the spatial details of the feature map~\cite{li2018harmonious,linsley2018global}. And some researcher even design a 3D-like attention module to extract the channel descriptors with spatial information~\cite{hu2018gather}. However, these methods bring a lot of additional parameters and computational consumption. Therefore, it is expected that the attention mechanism is paid attention to the details of feature maps under the framework of the channel-wise attention mechanism but with only a little additional cost.

This paper attempts to improve the representation of descriptors extracted by GAP through providing the feature maps with higher quality. The proposed method is called as Stochastic Region Pooling (SRP), which does not brings extra parameters and computation in test phase. SRP emphasizes more local features in the convolutional layer, making the channel descriptors extracted by GAP more representative and thus making the channel-wise attention mechanism works better. In more details, it stochastically selects the region from the feature map and used GAP to obtain the region descriptor, where the descriptor is the accurate representation of the region. Subsequently, the region descriptors are used in the follow-up attention structure. In such case, the back propagation~\cite{lecun1989backpropagation} will encourage these regions to have more important feature responses to represent its original entire feature map.

This paper proposes a simple method to implement SRP, named as Single Square SRP (SS-SRP), which stochastically selects a single square region from feature map to extract the descriptor. In order to consider local response of irregular shape in feature map, another method named as  Multiple Squares SRP (MS-SRP) is proposed that stochastically selects multiple square regions from the feature map and then extracts the descriptor from their union regions. These two methods are illustrated in figure.\ref{fig:1}. On the other hand, in residual networks, most of attention mechanisms only act on the residual branch of residual block. In order to make the feature maps of identity branch also have more feature responses, we use SRP to extract the channel descriptors of both identity branch and residual branch, which are then applied to serve the follow-up attention structure. The main contributions are as follows:
\begin{figure*}
\centering 
  \includegraphics[width=16cm]{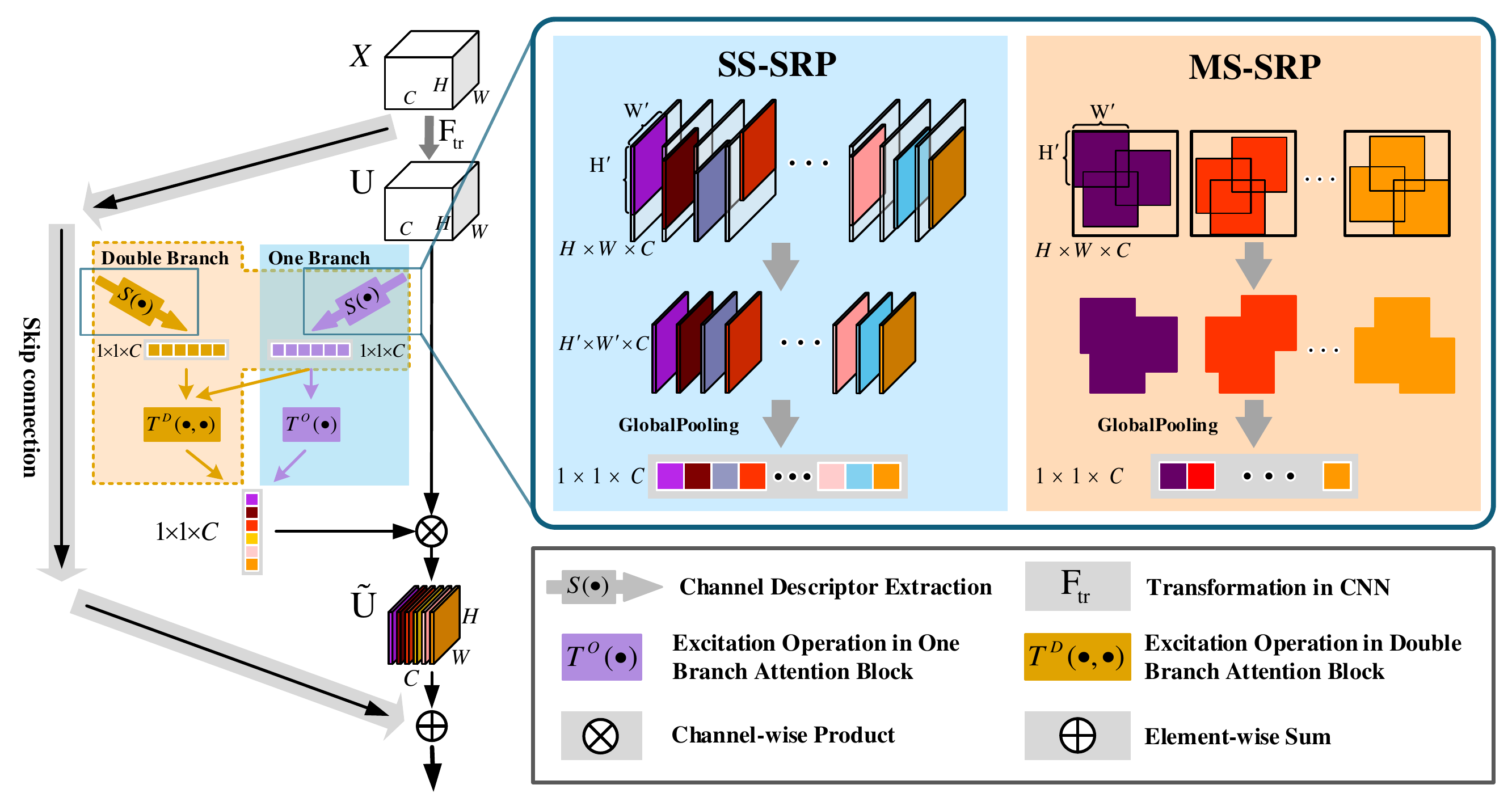}
  \caption{The framework of Stochastic Region Pooling (SRP) with applications to one branch or double branch attention block, where SS-SRP and MS-SRP are considered.}
  \label{fig:1}
\end{figure*}
\begin{itemize}
\item A new method SRP is proposed that make more local regions in the feature map to have more important feature responses, so that the channel descriptors extracted later by GAP are more representative and robust.
\item SRP obtains the significant improvement of the performance for one branch or double branch block of channel-wise attention structures without any additional parameters and computation in test phase. It can also work well with some augmentation methods to further improve the performance.
\item A linear strategy is proposed to make SRP work well that gradually reduces the scale ratio of region as the depth of the layer increases.
\item Experiments are conducted on serval datasets, including CIFAR-10/100, ImageNet and three Fine-grained datasets (CUB-200-2011, Stanford Cars and Stanford Dogs), which verified the effectiveness of SRP.
\end{itemize}




\begin{algorithm*}
\KwIn{Feature map:$U$, height and width of the feature map: $H$ and$W$, scale ratio:$\lambda$, the number of square regions:$M$, $mode$. }
\KwOut{The channel descriptors $z$ ($z_c$ is the $c^{th}$ channel descriptor of $z$).}
\If{$mode$ != $Training Stage$}{
$\forall c : z_c=F_{sq}(u_c) \qquad \qquad \qquad \qquad \rhd {F}_{sq}$ computes Eq.~\eqref{eqn:01}\;
	return $z$\;
	}
Calculate the height and width of square region: $H'=\lfloor \lambda H + \frac{1}{2}\rfloor, W'=\lfloor \lambda W + \frac{1}{2}\rfloor$\;

Stochastically sample $M$ positions $P^m_{i,j}$ from the feature map $U$ where $1\leq i \leq H-H'+1$, $1\leq j \leq W-W'+1$\;

Crop $M$ square regions from the feature map $U$ with the left-top position as $P^m_{i,j}$, where the width and height of square region are $W'$ and $H'$\ respectively;
$\forall c: {z}_c={F^{\*}}_{sq}(u_c,P_{i,j},H',W') \qquad \qquad  \rhd {F^{\*}}_{sq}$ computes Eq.~\eqref{eqn:04} or Eq.~\eqref{eqn:06}\;
return ${z}$\ ;
\caption{Stochastic Region Pooling}
\label{alg:one}
\end{algorithm*}

\section{Related Work}

\noindent\textbf{Improving representation of feature maps}.$\quad$ A common way to obtain high-quality feature maps is to find efficient network structures, such as~\cite{simonyan2014very, szegedy2015going, he2016deep,huang2017densely,yang2018convolutional} to extract more and better features. However, the feature maps learned by these networks are still not diverse enough. Another way is regularization. Some regularization for channel can maintain high quality channels by removing or retraining the inefficient channels ~\cite{he2017channel,zhang2016picking,gao2018dynamic,hou2019Weighted}. For ~\cite{he2017channel,zhang2016picking}, they will change the network structure. And for ~\cite{gao2018dynamic,hou2019Weighted}, we have not found the evidence to prove that they are suitable for channel-wise attention neural networks. Other regularizations such as dropout~\cite{srivastava2014dropout}, droppath~\cite{larsson2017fractalnet}, dropblock~\cite{ghiasi2018dropblock}, cutout~\cite{devries2017improved} can enhance the robustness of the feature by introducing randomness. And our method is closely related to Dropblock~\cite{ghiasi2018dropblock} which drops spatially correlated information to promote the network to reconstruct the important features from its surrounding. 
However, our method is aims to solve the problem that the descriptor in the channel-wise attention network contains few detailed information of the feature map, such as by promoting the feature map to have more or wider important feature responses.

\noindent\textbf{Extracting descriptors by spatial feature pooling}.$\quad$ The idea of spatial feature pooling was proposed by Hubel and Wiesel~\cite{hubel1962receptive}, and then Yann Lecun~\cite{lecun1998gradient} successfully applied it to CNN. Furthermore, Spatial Pyramid Matching(SPM)~\cite{lazebnik2006beyond,yang2009linear} manually designed the pooling weights to obtain spatial feature pyramid,  Malinowski~\cite{malinowski2013learning} parameterized pooling operator to learn the pooling regions, and Lee~\cite{lee2016generalizing} combined the max pooling and the average pooling to obtains a generalized pooling function. Some researchers also use the second-order pooling even the third-order pooling instead of the first-order pooling (i.e., GAP) to collect richer statistics of the last convolution layer in CNN~\cite{cui2017kernel,lin2015bilinear,wang2018global}. Introducing randomness can also improve the performance of spatial pooling. For example, stochastic pooling~\cite{zeiler2013stochastic} randomly selects the activation value based on a multinomial distribution formed by activations of each pooling region to regularize the network, and S3Pool~\cite{zhai2017s3pool} randomly picks feature map's rows and columns and then performs the max pooling operation to implicitly introduce data augmentation. However, using the above method to extract the channel descriptor will bring a lot of extra consumption, or it will still not make the descriptor's representation stronger. We select a simplest way, which takes GAP to extract the descriptor because it is widely used, does not bring any extra parameters, and has the potential to get global spatial information.

\noindent\textbf{Methods of spatial pooling in attention mechanism}.$\quad$ Channel-wise attention mechanisms have developed rapidly in recent years~\cite{nguyen2018attentive,chen2017sca,li2018harmonious,hu2018squeeze,hu2018gather,woo2018cbam} and the channel descriptors are crucial to them.
In order to extract more representative descriptors, CBAM~\cite{woo2018cbam} combines the output of the global max pooling and the global average pooling as the pooling method, and GEnet~\cite{hu2018gather} uses the depth-wise convolution with large kernel to replace GAP. 
However, the global max pooling in CBAM is prone to network overfitting~\cite{zeiler2013stochastic}, and CBAM also cannot enhance the channel descriptor with more spatial details. Besides, GENet will brings a lot of additional parameters or computation.
Different from them, SRP is a training method that can encourage the descriptors to have more information about the feature map details. The reason why SRP uses GAP instead of the above methods to extract the descriptor is not only because GAP is simple and does not bring any additional parameters, but also that GAP is widely used for attention networks. This enables SRP to be conveniently used on these networks without modifying the network structure.


\section{Stochastic Region Pooling}
Many channel-wise attention block applied GAP operation to obtain the descriptor of feature map. Formally, given the feature maps $U=[u_1,u_2,\cdots,u_c]\in \mathbb{R}^{H\times W\times C}$ in the convolutional layer and the function $S(\cdot)$ squeezes the global spatial information into a channel descriptor $z = [z_1,z_2,\cdots,z_C]\in R^{C}$, the $c^{th}$ channel descriptor of $z$ can be calculated in GAP by
\begin{equation}
\label{eqn:01}
z_c=S(u_c)=\frac{1}{H\times W}\sum^{H}_{i=1}\sum^{W}_{j=1}u_c(i,j).
\end{equation}

From the Eq.\ref{eqn:01}, it can be concluded that GAP regards each position of the space to make the same contribution, even if the elements of some local regions have the low magnitude~\cite{zeiler2013stochastic}. This will weaken the details of the feature map and easily leads to high similarity between descriptors. Here we propose Stochastic Region Pooling (SRP) method that stochastically selects the region from the feature map instead of the whole map to extract descriptors during the training stage. SRP is presented as Algorithm ~\ref{alg:one}.


\textbf{SS-SRP} is a simple method to be implemented, which stochastically selects a single square region from map. Supposing that the scale ratio $\lambda$ controls the size of square region, the width and height of the square region can be formulated as follows,
\begin{equation}
\label{eqn:03}
H'=\lfloor \lambda H + \frac{1}{2}\rfloor, W'=\lfloor \lambda W + \frac{1}{2}\rfloor.
\end{equation}
For each feature map, we stochastically select a position $P(a,b)$ as the upper left corner of the square region $R\in \mathbb{R}^{H'\times W'}$, where the spatial position $P(a,b)$ subject to $1\leq i \leq H-H'+1$, $1\leq j \leq W-W'+1$.
Now we use the average pooling as the squeeze operator $S(\cdot)$ to extract the descriptors $z$, and the $c^{th}$ channel descriptor of $z$ can be calculated by 
\begin{equation}
\label{eqn:04}
{z}_c=S(u_c)=\frac{1}{H'\times W'}\sum^{a+H'-1}_{i=a}\sum^{b+W'-1}_{j=b}u_c(i,j).
\end{equation}
The module of SS-SRP can be seen in the Figure \ref{fig:1}.

\textbf{MS-SRP} is applied to consider the non-regular shape of the local region in feature map, which stochastically selects multiple square regions from the feature map and then extracts the descriptor from their union regions. Suppose that we stochastically choose $M$ square regions from the feature map, defined as  $R \in \mathbb{R}^{M\times H\times W}$, the target region we want is their union area $R^{*}$.

Let $\Omega^{*}$ be the set of all the points in $R^{*}$ and $\Omega_m$ be the set of all points in the $m^{th}$ square region $R_m$, we have
\begin{equation}
\label{eqn:05}
\Omega^{*}= \bigcup_{m = 1}^{M} \Omega_m=\bigcup_{m = 1}^{M} \big\{(x,y)\mid(x,y)\in \Omega_m\big\}.
\end{equation}

The global average pooling is used as we did before to squeeze the regional spatial information. Thus the $c^{th}$ channel descriptor of $z$ can be calculated by 
\begin{equation}
\label{eqn:06}
{z}_c=S(u_c,\Omega^{*})=\frac{1}{\mid\Omega^{*}\mid}\sum_{(i,j)\in \Omega^{*}}u_c(i,j),
\end{equation}
where $\mid\Omega^{*}\mid$ is the number of elements in $\Omega^{*}$,i.e. the number of points in region $R^{*}$. The module of MS-SRP can be seen in the Figure \ref{fig:1}.


\noindent\textbf{One branch or double branch attention block.}
\label{sec:3}
After calculating the channel descriptors, a one branch attention block applies an excitation operation $T^{O}(\cdot)$ to obtain the relationship $\mathbb{\alpha}\in \mathbb{R}^{C}$ between the channels of the residual branch, which is $\alpha=T^{O}\big(z^r)$ where $z^r$ is the channel descriptor of the residual branch. And a double branch attention block utilizes $T^{D}(\cdot,\cdot)$ to compute the relationship among channels of the residual branch, which is $\alpha=T^{D}\big(z^{id},z^{r})$, where $z^{id}$ and $z^{r}$ are the channel descriptors of the identity branch and the residual branch respectively.

For the one branch attention block, we use two fully connected(FC) layers as the function $T^{O}\big(z^r)$ as described in SENet~\cite{hu2018squeeze}. For the double branch attention block, we fold two branch's descriptors and use convolution $3\times 3$ to model their relationship as the function $T^{D}\big(z^{id},z^{r})$ as described in CMPE-SENet~\cite{hu2018competitive}. 

Finally, these recalibrated feature maps $\widetilde{U}\in \mathbb{R}^{H\times W \times C}$ can be calculated by $\widetilde{U}= \alpha \cdot U$, where $\cdot$ is the element-wise multiplication.


\noindent\textbf{Scheduled SRP.}$\quad$ The neurons in the shallow layers of the network have smaller receptive field. In order to maintain a majority of responses in the region selected by SRP in the shallow layers, we gradually reduce $\lambda$ from 1 to the smaller value as the depth of the layer increases, instead of setting $\lambda$ to a fixed value. In our experiments, we use linear strategy to reduce  $\lambda$, which is inspired by ScheduledDropPath~\cite{zoph2018learning}, but ScheduledDropPath changes its parameters over training time.


\section{Experiments}

Some experiments are conducted to validate the proposed method on the CIFAR-10/100\cite{krizhevsky2009learning}, ImageNet\cite{ILSVRC15}, and three fine-grained datasets (CUB-200-2011\cite{wah2011caltech}, Stanford Dogs\cite{khosla2011novel} and Stanford Cars\cite{krause20133d}). For experiments on CIFAR and fine-grained datasets, we report the average accuracy by running five times, while on ImageNet, we report the average accuracy by doing three times due to the limitation of computational resources. In the following subsections, SRP-O indicates that SRP is applied to the one branch attention block, and SRP-D indicates that SRP is applied to the double branch attention block.

\subsection{Experiment Settings}

\begin{table}
\begin{minipage}{\columnwidth}
\begin{center}
\scalebox{1}{
\begin{tabular}{lccc}
  \toprule
Datasets		& \#Class	& \#Train	& \#Test \\
  \midrule
CUB-200-2011	& 200		& 5,994		& 5,794	\\
Stanford Cars	& 196		& 8,144		& 8,041	\\
Stanford Dogs	& 120		& 12,000	& 8,580	\\
  \bottomrule
\end{tabular}}
\end{center}
\end{minipage}
\caption{Statistics	of three common Fine-grained datasets.}
\label{tab:1}
\end{table}%
\begin{figure}
\centering
\includegraphics[width=4cm]{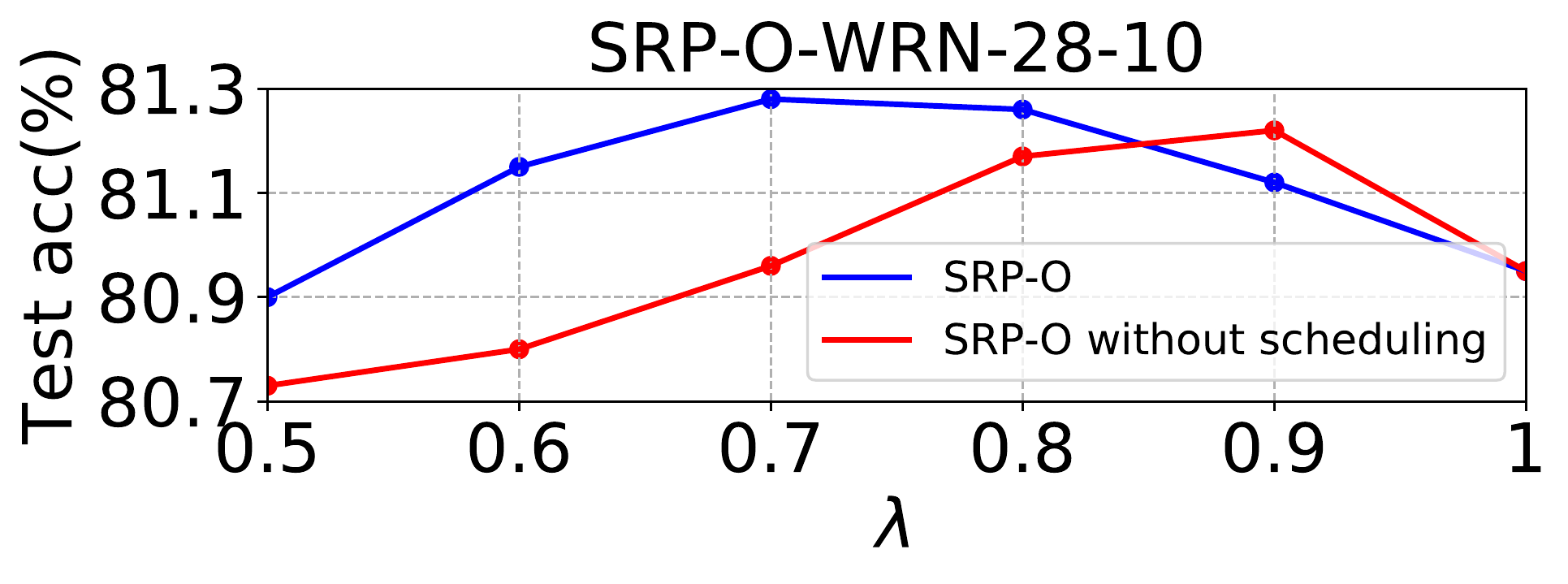}
\includegraphics[width=4cm]{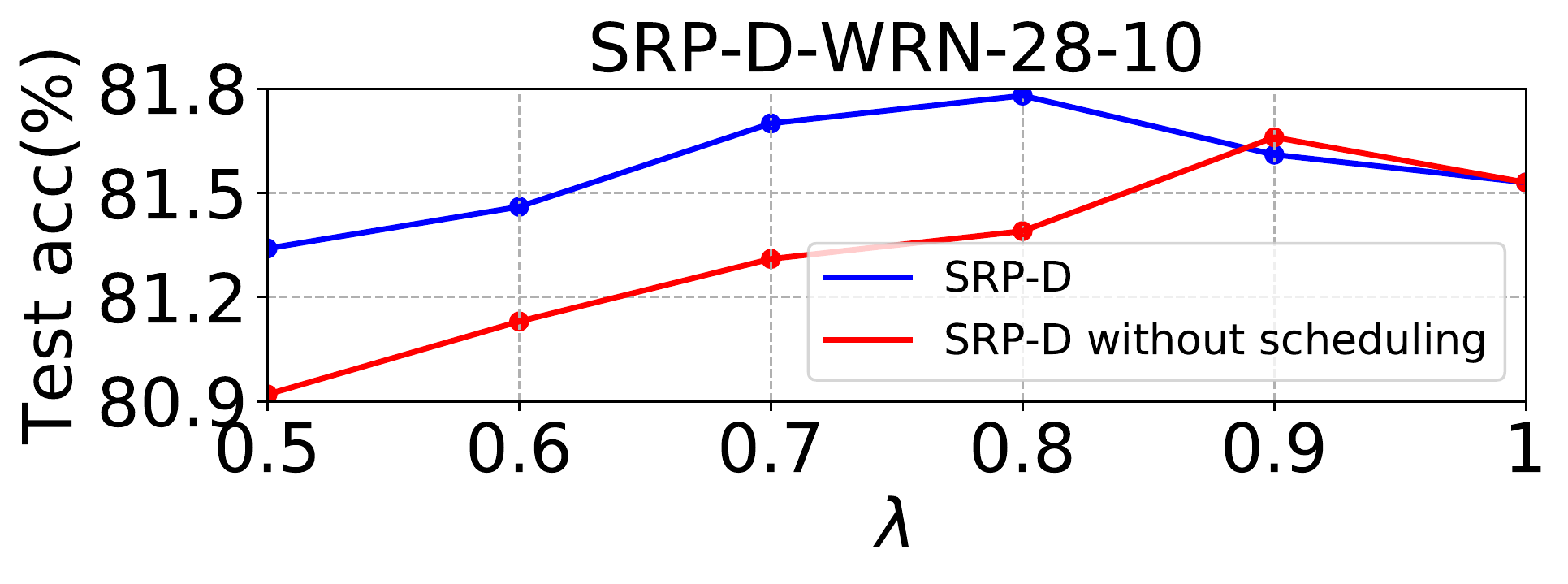}
\includegraphics[width=4cm]{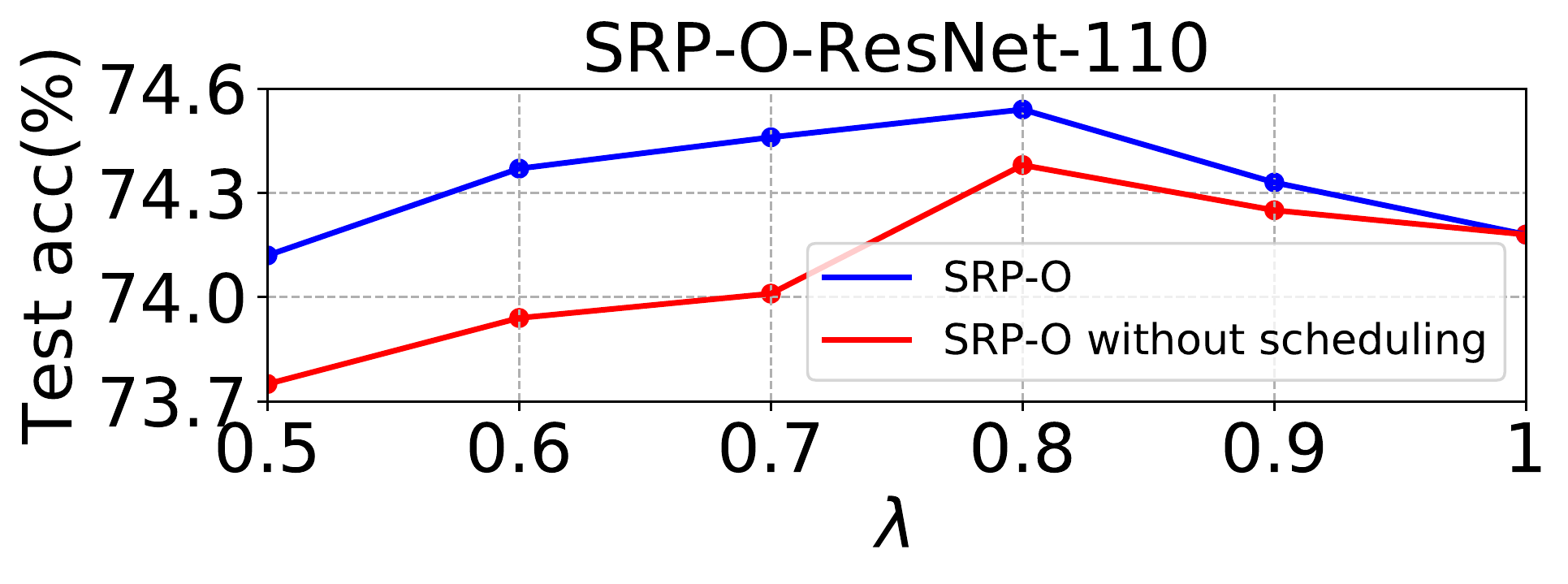}
\includegraphics[width=4cm]{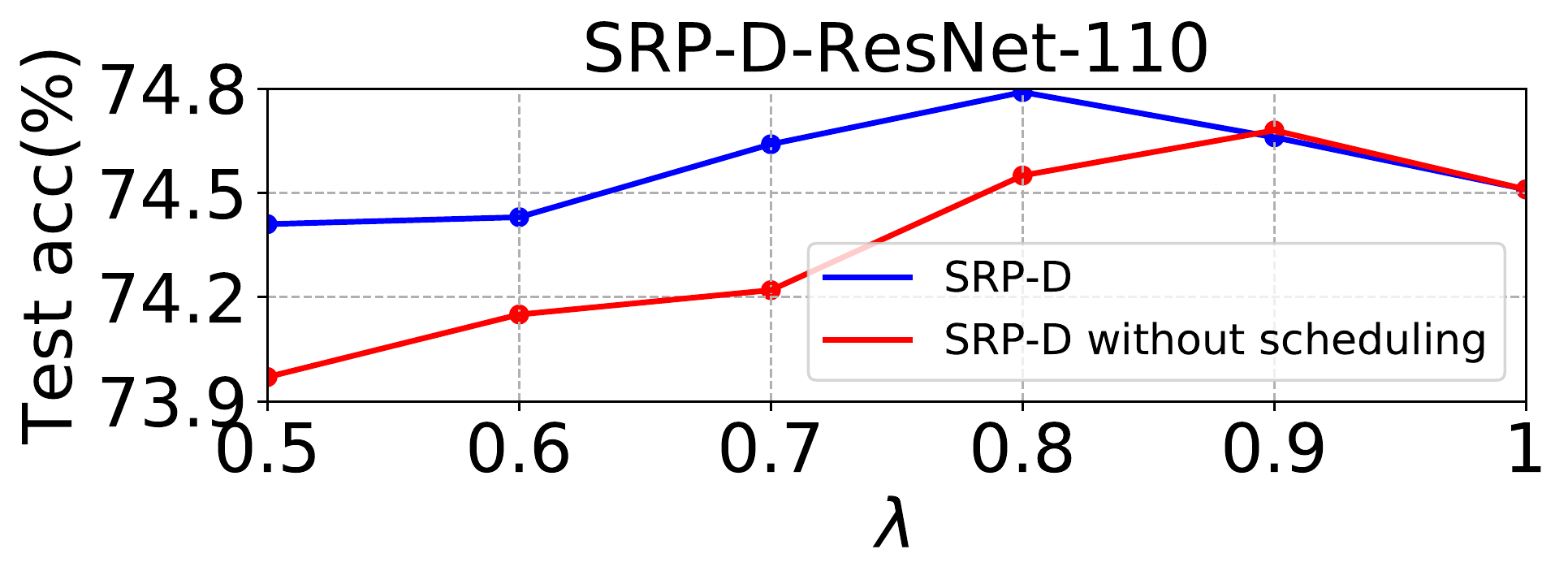}
\caption{Testing acc (\%) of SS-SRP (with or without scheduled) applied to One or Double branch block on CIFAR-100 data with the different $\lambda$.}
\label{fig:2}
\end{figure}

\begin{figure}
\centering
\includegraphics[width=4cm]{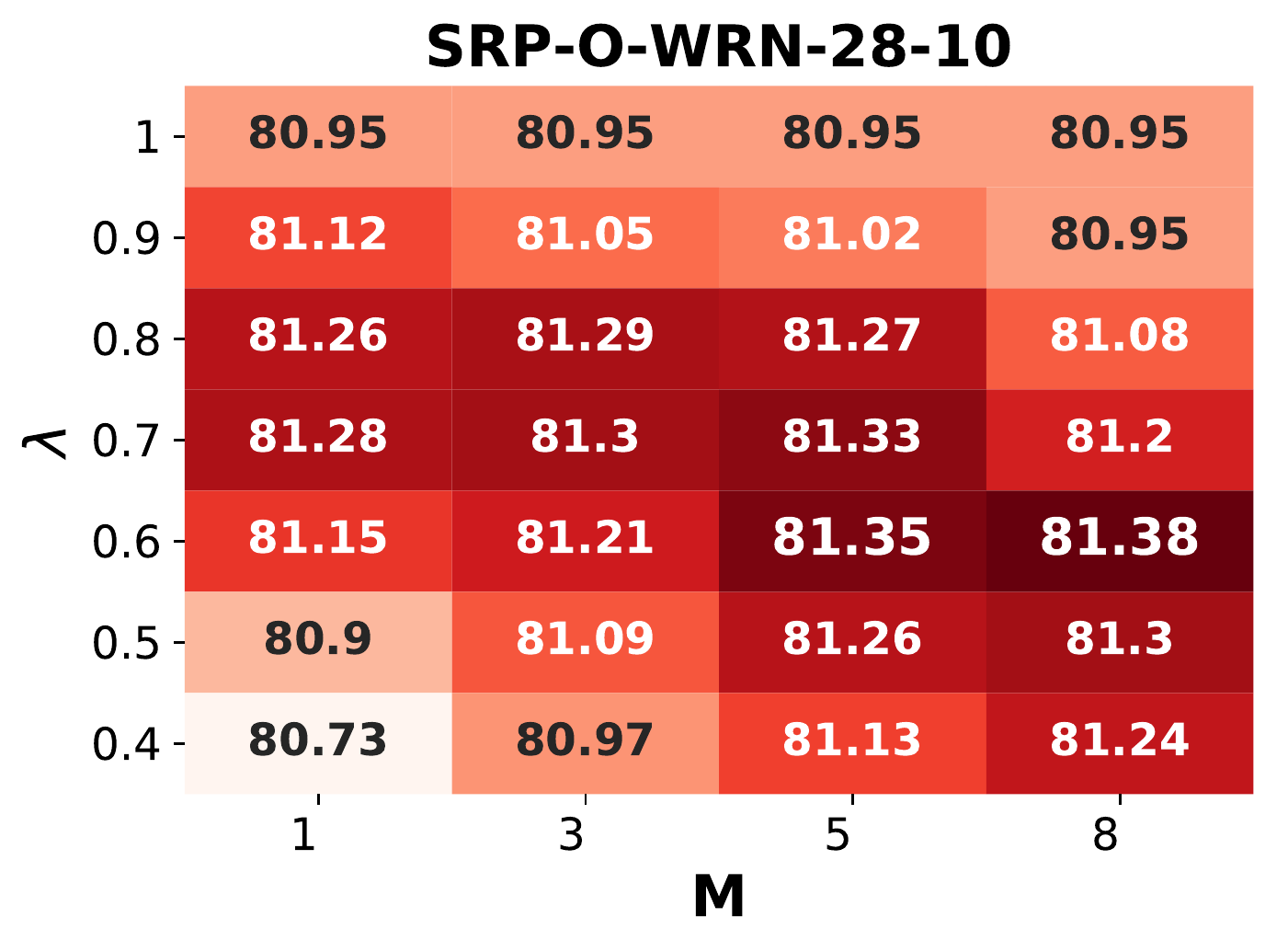}
\includegraphics[width=4cm]{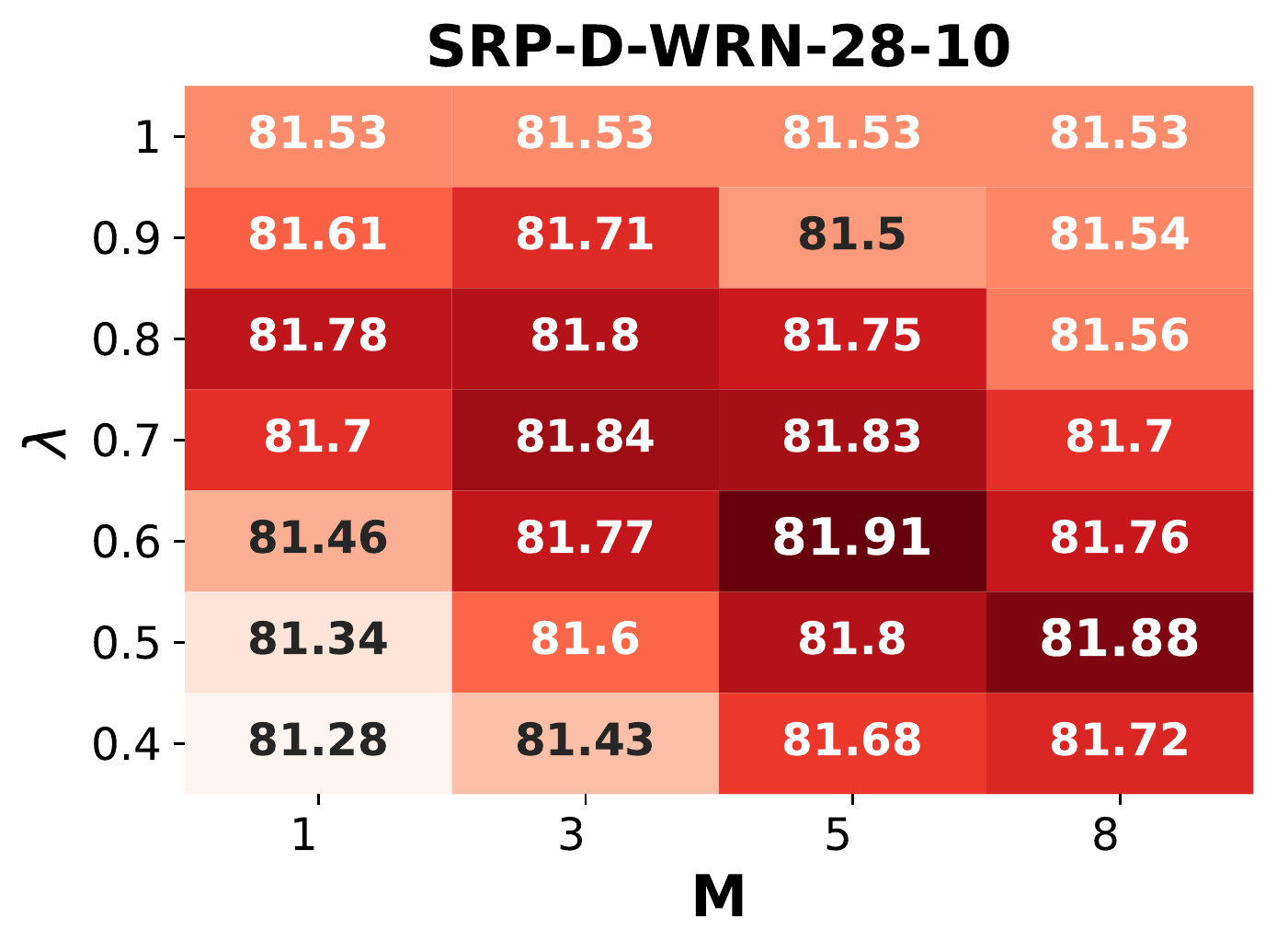}
\caption{Test acc (\%) of MS-SRP applied to One or Double Branch block on CIFAR-100 dataset under different hyperparameters ($\lambda, M$).}
\label{fig:3}
\end{figure}
\noindent\textbf{CIFAR.} Following ~\cite{he2016identity,Zagoruyko2016WRN,xie2017aggregated}, we use stochastic gradient descent (SGD) with 0.9 Nesterov momentum and batchsize of 128. The learning rate is set to be 0.1, which is then divided by 10 at epochs 100, 150 for ResNet, divided by 5 at epochs 60,120,160 for WRN, and divided by 10 at epochs 150,225 for ResNeXt. We train the model by 200 epochs for ResNet and WRN, and 300 epochs for ResNeXt. The weight decay is 0.0001 for ResNet, 0.0005 for WRN and ResNeXt. we use the standard data augmentation (translation/mirroring) for the training sets.

\noindent\textbf{ImageNet.} The ILSVRC 2012 contains 1.2 million training images and 50K validation images with 1K classes. We adopt the standard data augmentation for the training sets, which randomly samples a 224$\times$224 crop from the original images or their horizontal flip, and applies a single-crop with the size 224$\times$224 at testing stage. We train our models for 100 epochs and drop the learning rate by 0.1 at epoch 30, 60, and 90, and use SGD with the mini-batch size of 256 on 4 GPUs (64 each). The weight decay is 0.0001 and Nesterov momentum is 0.9. In experiments, we report the classification accuracy on the validation set.

\noindent\textbf{Fine-grained Datasets.} We conduct experiments on three fine-grained datasets, including CUB-200-2011,Stanford Dogs and Stanford Cars. The detailed statistics of each dataset are shown in Table \ref{tab:1}. For all fine-grained datasets, we resize the input images to 512$\times$512 and randomly crop the smaller images with 448$\times$448 from it, and then generates the horizontal flip of the cropped images for training. At the testing stage, we only use a single cropped image with 448$\times$448 from the input image which have been resized to 512$\times$512. We fine-tune networks (pre-trained on ImageNet) using SGD with the batch size of 16, momentum of 0.9 and weight decay of 0.00001. For all fine-grained datasets, we train the networks for 90 epochs. The learning rate begins with 0.001 and then divided by 10 at epoch 30 and 60.

\subsection{Impact of Hyper-parameters}

In order to demonstrate the influence of two hyper-parameters (scale ratio $\lambda$ and the square regions number $M$) on the performance of our model, experiments are conducted on CIFAR-100, where SRP applied on attention networks with different hyper-parameter settings.

\noindent\textbf{SS-SRP}. It only randomly selects one square region, which becomes standard mothod when we set $\lambda=1$. It can be seen from figure \ref{fig:2} that an appropriate $\lambda$ can improve the network performance and the lower $\lambda$ will result in the poor results. In the following experiments of SS-SRP, we use $\lambda=0.8$ unless specified elsewhere, because SS-SRP obtains the better results at $\lambda=0.8$.
From figure \ref{fig:2}, it can be also observed that SRP with the fixed scale ratio $\lambda$ can effectively improve the network performance, but the scheduled SRP makes the network work better. In all following experiments of SRP, scheduled SRP is used.


\noindent\textbf{MS-SRP}. In MS-SRP, there are two hyper-parameters that affects the network performance, i.e., the scale ratio $\lambda$ and the square number $M$. By observing the results in figure \ref{fig:3}, MS-SRP outperforms standard method SRP (i.e., $\lambda=1$) and SS-SRP within a wider range of hyper-parameters. Since $\lambda=0.6$ and $M=5$ can obtain the better results, we use these hyper-parameters for all experiments of MS-SRP unless stated elsewhere.



\begin{table}
\centering
\scalebox{0.75}{
\begin{tabular}{lcccccc}
  \toprule
  Model 																& depth 	& params	 & C10 			& C100	\\
  \midrule  	 						
  FractalNet~\cite{larsson2017fractalnet}								& 21		& 38.6M			& 95.40			& 76.27	\\
  WRN-28-10~\cite{Zagoruyko2016WRN}										& 28		& 36.5M			& 96.00			& 80.75 \\   		
  ResNeXt-29(8x64d)~\cite{xie2017aggregated}						    	& 29		& 34.4M			& 96.35 			& 82.23 \\
  ResNeXt-29(16x64d)~\cite{xie2017aggregated}						    & 29		& 68.1M			& 96.42 			& 82.69 \\
  DenseNet(k=24)~\cite{huang2017densely}									& 100		& 27.2M			& 96.26 			& 80.75	\\
  DenseNet-BC(k=40)~\cite{huang2017densely}								& 190		& 25.6M			& 96.54			& 82.82	\\
  PyramidNet(bottleneck,$\alpha=270$)~\cite{han2017deep}					& 272		& 27.0M			& 96.52			& {\color{red}{82.99}}	\\
  \midrule  							
  \textit{mixup}~\cite{zhang2017mixup}, WRN-28-10~\cite{Zagoruyko2016WRN} & 28		& 36.5M			& 97.30		& 82.50 \\						
  \textit{mixup}~\cite{zhang2017mixup}, DenseNet-BC(k=40)~\cite{huang2017densely} & 190	& 25.6M			& 97.30			& 83.20 \\				
  \textit{mixup}~\cite{zhang2017mixup}, SE-WRN-28-10~\cite{hu2018squeeze} & 28		& 36.8M			& 97.32			& 83.23 \\	
  \midrule  	 						
  \textbf{SS-SRP-O}, WRN-28-10 							& 28		& 36.8M			& 96.28			& 81.38		\\
  \textbf{SS-SRP-O}, ResNeXt-29(8x64d) 					& 29		& 34.9M			& 96.52			& 82.59	\\
  \textbf{SS-SRP-D}, WRN-16-8 							& 16		& 11.1M			& 96.02			& 80.89		\\
  \textbf{SS-SRP-D}, WRN-28-10 							& 28		& 36.9M			& 96.50			& 81.78		\\
  \midrule
  \textbf{MS-SRP-O}, WRN-28-10 							& 28		& 36.8M			& 96.34			& 81.38		\\
  \textbf{MS-SRP-O}, ResNeXt-29(8x64d) 					& 29		& 34.9M			& 96.53			& \textbf{82.62}	\\
  \textbf{MS-SRP-D}, WRN-16-8 							& 16		& 11.1M			& 96.10 	 	& 80.98	\\
  \textbf{MS-SRP-D}, WRN-28-10 							& 28		& 36.9M			& {\color{red}{96.61}} & 81.91		\\
  \midrule  									
  \textbf{MS-SRP-O}, \textit{mixup}, WRN-28-10 			& 28		& 36.8M			&  97.48				& 84.08	 	 	\\
  \textbf{MS-SRP-D}, \textit{mixup}, WRN-16-8 			& 16		& 11.1M			&  96.84				& 82.71 	 	\\
  \textbf{MS-SRP-D}, \textit{mixup}, WRN-28-10 			& 28		& 36.9M			& {\color{red}{97.56}}	 & {\color{red}{84.12}}	 	\\
  \bottomrule
\end{tabular}}
\caption{Comparison of test accuracy (\%) with different methods on the CIFAR-10 and CIFAR-100.  The best results are highlighted in {\color{red}{red}}, and the best records of our models are in \textbf{bold}. Combined with the augmentation method of \textit{mixup}~\cite{zhang2017mixup}, SRP can challenge state-of-the-art results.}
\label{tab:2}
\end{table}%

%

\begin{table}
\begin{minipage}{\columnwidth}
\begin{center}
\scalebox{0.8}{
\begin{tabular}{lccc }
  \toprule
  Model 														& params  	& top-1		& top-5  \\
  \midrule
  CliqueNet-S3~\cite{yang2018convolutional}						& 14.4M 	& 75.95		& 92.85	\\
  ResNet-50~\cite{he2016deep}										& 25.6M		& 75.30		& 92.20	\\
  ResNet-101~\cite{he2016deep}									& 44.6M		& 76.40		& 92.90	\\
  SE-ResNet-50~\cite{hu2018squeeze}								& 28.1M		& 76.71		& 93.38	\\
  ResNet-152~\cite{he2016deep}									& 28.1M		& 77.00		& 93.30	\\
  DenseNet-201~\cite{huang2017densely}							& 20.2M		& 77.42      & 93.66 \\
  SE-ResNet-101~\cite{hu2018squeeze}								& 49.4M		& 77.62		& 93.93	\\
  CBAM-ResNet-50~\cite{woo2018cbam}								& 25.9M		& 77.34		& 93.69	\\
  GE-$\theta^+$-ResNet-50~\cite{hu2018gather}						& 33.7M		& \textbf{78.12} & 94.20	\\
  \midrule
  \textbf{SS-SRP-O}-ResNet-50									& 28.1M		& 77.43		& 93.81	\\
  \textbf{MS-SRP-O}-ResNet-50									& 28.1M		& 77.58		& 93.88	\\
  \textbf{SS-SRP-D}-ResNet-50									& 29.2M		& 77.94		& 94.35	\\
  \textbf{MS-SRP-D}-ResNet-50									& 29.2M		& 78.09		& \textbf{94.40}	\\
  \bottomrule
\end{tabular}}
\end{center}
\end{minipage}
\caption{Comparison of test accuracy (\%) between SRP and other different methods on the large ImageNet, where a single crop method is applied.}
\label{tab:3}
\end{table}%


\begin{table*}
\centering
\begin{minipage}[b]{0.7\columnwidth}
\scalebox{0.7}{
\begin{tabular}{lccc }
  \toprule
  Model 												& Anno.		 & 1-Stage		& Acc. 		\\
  \midrule
  DVAN~\cite{zhao2017diversified}						& $\times$ 	 & $\times$		& 79.0		\\
  Part-RCNN~\cite{zhang2014part}							& \checkmark & $\times$		& 81.6 		\\
  PA-CNN~\cite{krause2015fine}							& \checkmark & \checkmark	& 82.8 		\\
  RAN~\cite{wang2017residual}							& $\times$	 & $\times$		& 82.8 		\\
  FCAN~\cite{liu2016fully}								& \checkmark & \checkmark	& 84.7 		\\
  RACNN~\cite{fu2017look}								& $\times$	 & $\times$		& 85.3 		\\
  \midrule
  VGG-19~\cite{simonyan2014very}							& $\times$ 	 & \checkmark	& 77.8		\\
  ResNet-50~\cite{he2016deep}							& $\times$   & \checkmark	& 81.7  	\\
  DenseNet-161~\cite{huang2017densely}					& $\times$   & \checkmark	& 84.2  	\\
  FCAN~\cite{liu2016fully}								& $\times$	 & \checkmark	& 84.3 		\\
  ResNet-101~\cite{he2016deep}							& $\times$   & \checkmark	& 84.5  	\\
  \midrule
  \textbf{SS-SRP-D}-ResNet50						& $\times$	 & \checkmark	& 84.9		\\
  \textbf{MS-SRP-D}-ResNet50						& $\times$	 & \checkmark	& \textbf{85.6}		\\
  \bottomrule
\end{tabular}}
\subcaption{CUB-200-2011.}
\label{tab:4a}
\end{minipage}\begin{minipage}[b]{0.7\columnwidth}
\scalebox{0.7}{
\begin{tabular}{lccc }
  \toprule
  Model 												& Anno.		 & 1-Stage		& Acc. 		\\
  \midrule
  DVAN~\cite{zhao2017diversified}													& $\times$ 	 & $\times$		& 87.1		\\
  RAN~\cite{wang2017residual}										 			& $\times$	 & $\times$		& 91.0 		\\
  FCAN~\cite{liu2016fully}													& \checkmark & \checkmark	& 91.3 		\\
  RACNN~\cite{fu2017look}													& $\times$   & $\times$		& 92.5  	\\
  PA-CNN~\cite{krause2015fine}												& \checkmark & \checkmark	& \textbf{92.8} 		\\
  \midrule
  VGG-19~\cite{simonyan2014very}												& $\times$ 	 & \checkmark	& 84.9		\\
  FCAN~\cite{liu2016fully}													& $\times$	 & \checkmark	& 89.1 		\\
  ResNet-50~\cite{he2016deep}													& $\times$   & \checkmark	& 89.8  	\\
  DenseNet-161~\cite{huang2017densely}												& $\times$   & \checkmark	& 91.8  	\\
  ResNet-110~\cite{he2016deep}												& $\times$   & \checkmark	& 91.9  	\\
  \midrule
  \textbf{SS-SRP-D}-ResNet50						& $\times$	 & \checkmark	& 92.3		\\
  \textbf{MS-SRP-D}-ResNet50						& $\times$	 & \checkmark	& \textbf{92.8}		\\
  \bottomrule
\end{tabular}}
\subcaption{Stanford Cars.}
\label{tab:4b}
\end{minipage}\begin{minipage}[b]{0.7\columnwidth}
\scalebox{0.7}{
\begin{tabular}{lccc }
  \toprule
  Model 												& Anno.		 & 1-Stage		& Acc. 		\\
  \midrule
   DVAN~\cite{zhao2017diversified}													& $\times$ 	 & $\times$		& 81.5		\\
  RAN~\cite{wang2017residual}										 			& $\times$	 & $\times$		& 83.1 		\\
  \midrule
  VGG-16~\cite{simonyan2014very}												& $\times$ 	 & \checkmark	& 76.7		\\
  ResNet-50~\cite{he2016deep}												& $\times$   & \checkmark	& 81.1  		\\
  DenseNet-161~\cite{huang2017densely}											& $\times$   & \checkmark	& 81.2  		\\
  FCAN~\cite{liu2016fully}													& $\times$	 & \checkmark	& 84.2 		\\
  MAMC-ResNet50~\cite{sun2018multi}  										& $\times$	 & \checkmark	& 84.8 		\\
  ResNet-101~\cite{he2016deep}												& $\times$   & \checkmark	& 84.9  		\\
  \midrule
  \textbf{SS-SRP-D}-ResNet50					& $\times$	 & \checkmark	& 85.9			\\
  \textbf{MS-SRP-D}-ResNet50					& $\times$	 & \checkmark	& \textbf{86.3}		\\
  \bottomrule
\end{tabular}}
\subcaption{Stanford Dogs.}
\label{tab:4c}
\end{minipage}
\caption{Comparison results on three Fine-grained datasets including CUB-200-2011, Stanford Cars and Stanford Dogs. "Anno." represents using extra annotation in training. "1-Stage" represents whether the training can be done in one stage. "Acc." represents the test set accuracy (\%)}
\label{tab:4}
\end{table*}%

\begin{figure}
\centering
  \includegraphics[width=8cm]{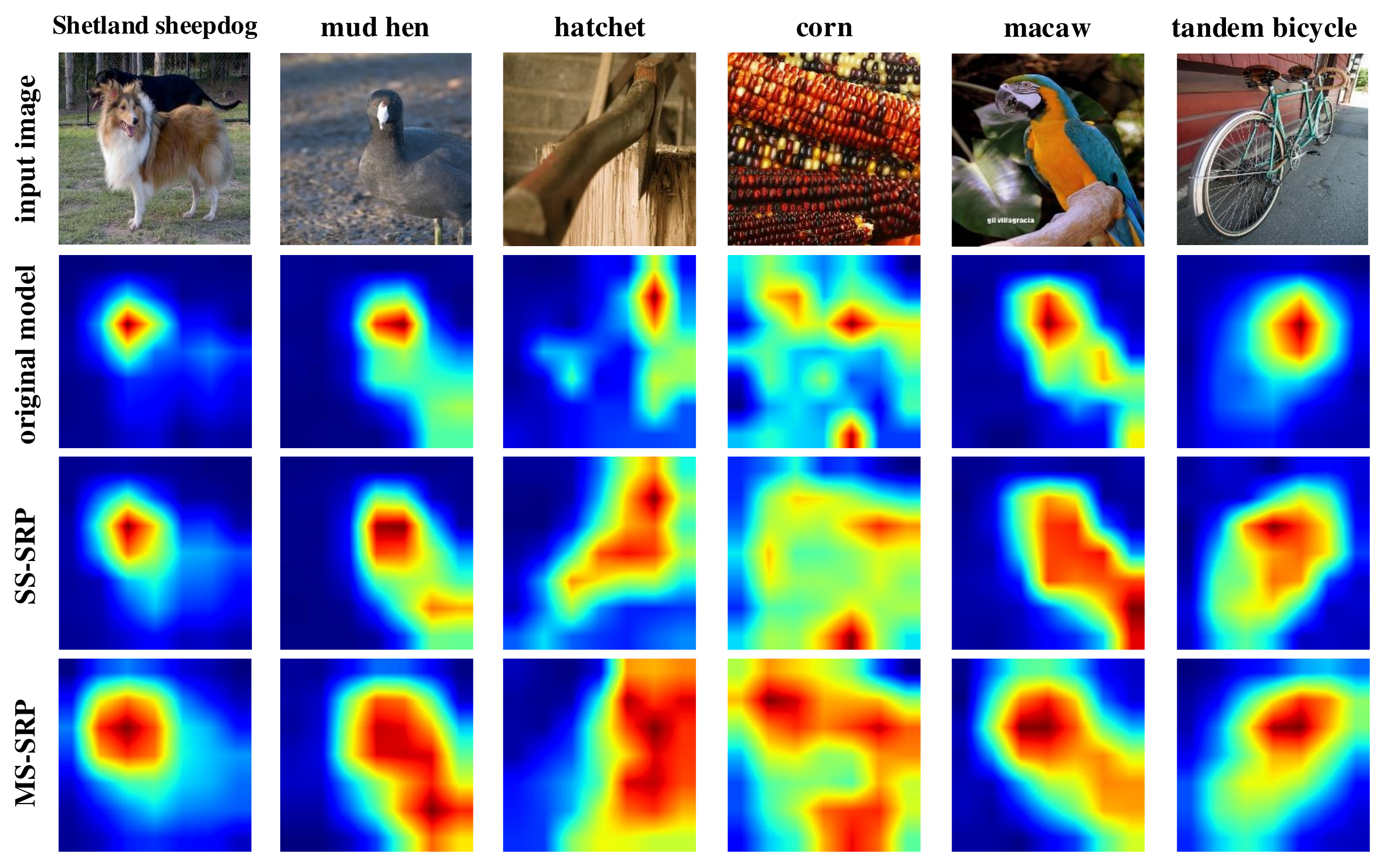}
  \caption{In \textbf{ImageNet}, the Grad-CAM~\cite{selvaraju2017grad} visualization for double branch block of attention ResNet50 model trained without SRP and trained with SRP. Best viewed in color.}
  \label{fig:4}
\end{figure}


\subsection{CIFAR Classification}

Table \ref{tab:2} presents the results of SRP and compared state-of-the-art CNN architectures on CIFAR. It can be observed that SRP method consistently achieve the better effective performance when applied it to other networks. Furthermore, the small network trained with SRP can achieve the comparable accuracy to some larger models, such as MS-SRP-O-ResNext-29 (34.9M) vs ResNext-29 (68.1M)~\cite{xie2017aggregated}, MS-SRP-D-WRN-16-8 (11.1M) vs DenseNet (27.2M)~\cite{huang2017densely} or WRN-28-10 (36.5M)~\cite{Zagoruyko2016WRN}. On the other hand, SRP may further improve network performance when augmentation methods are applied, such as \textit{mixup}~\cite{zhang2017mixup}. Combined with \textit{mixup} in the training stage, MS-SRP-D surpasses all the comparison methods, and can challenge state-of-the-art results.


\subsection{ImageNet Classification}

We next investigate the effectiveness of SRP on large dataset, the  ILSVRC 2012 dataset.

\noindent\textbf{Comparison with state-of-the-arts CNNs.}
It can be observed from Table \ref{tab:3} that SRP can still improve the network performance effectively on the large data sets and achieved very competitive accuracy. For example, it exceeds all the methods in terms of top-5 accuracy, while it is much close to the state-of-the-art method by top-1. For SE-ResNet-50, the accuracy improvement of SRP is 0.87\% on top-1 and 0.5\% on top-5. And the best result of SRP surpasses the basic ResNet-50 and SE-ResNet-50 more than 2\% and 1\% respectively, by both top-1 and top-5. Compared with the GE-$\theta^+$~\cite{hu2018gather}, SRP can achieve comparable or better performance but obviously uses fewer parameters. Moreover, compared with the counterparts, SRP can achieve higher accuracy by top-5 while their accuracy are similar by top-1. This is because SRP promotes the feature maps to contain more features response, making the network have better generalization ability.

%

\noindent\textbf{SRP learns more and wider regions.} The model trained with SRP will promote more local feature responses in the convolutional layer, making the network focus on the more and wider regions. Here, we use Grad-CAM~\cite{selvaraju2017grad} to visualize the importance of the spatial position in the convolutional layer. The visualization results of SRP networks (SS-SRP-D-ResNet50, MS-SRP-D-ResNet50) and the baseline (ResNet50 with double branch attention block) can been seen in Figure \ref{fig:4}. It can be clearly observed that the Grad-CAM masks of SRP cover the target object better and wider than the baseline. That is, the model trained with SRP tends to focus on several spatially distributed regions, and aggregate information from multiple or wider regions.

\begin{figure*}
\centering
  \includegraphics[width=16cm]{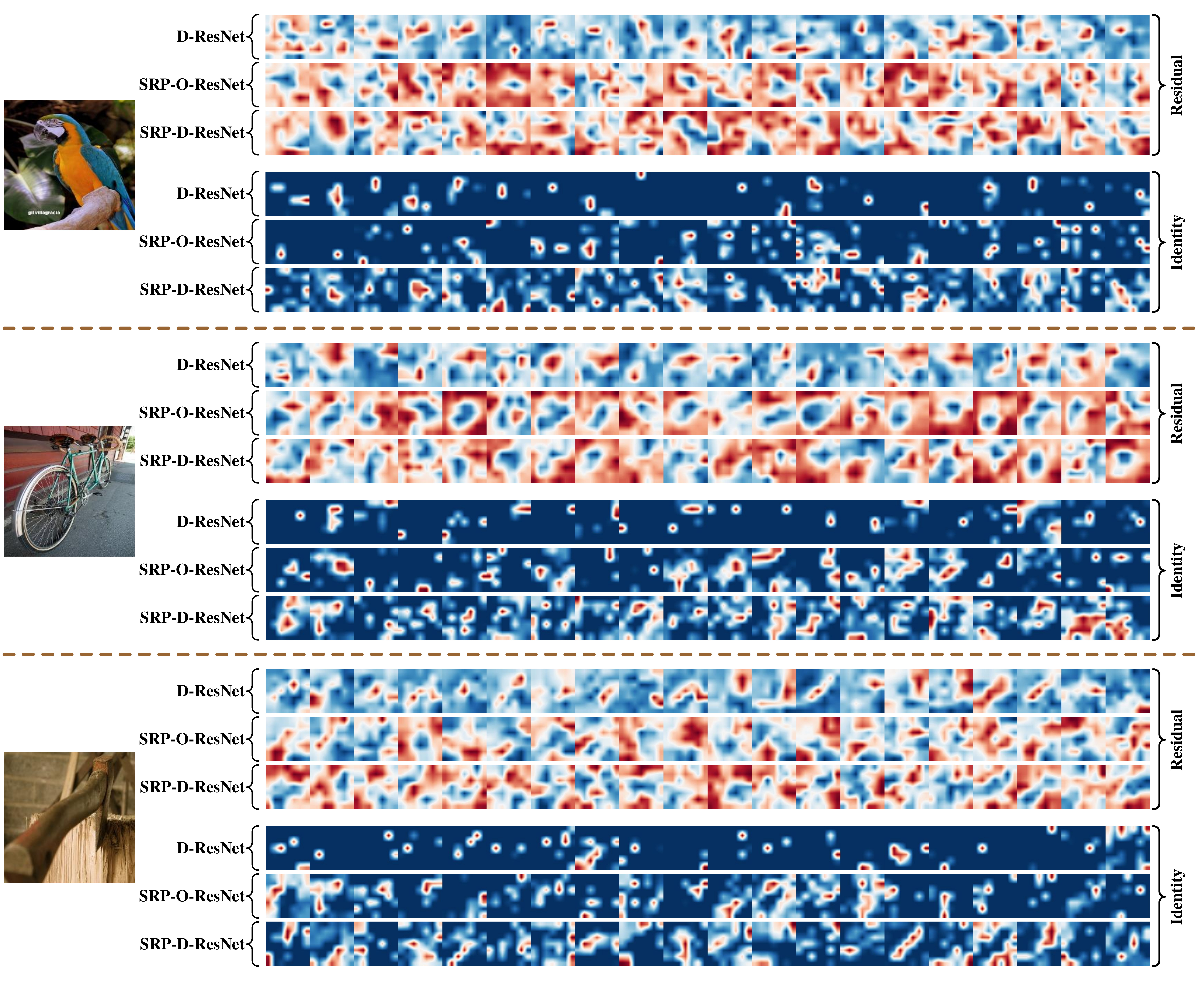}
  \caption{The feature maps of ResNet-50 with different mode. D-ResNet means ResNet-50 with double branch attention block. SRP-O-ResNet and SRP-D-ResNet means ResNet-50 with one or double branch attention block and train with MS-SRP. These feature maps are from the residual branch or identity branch of the 14th block of the network. We only display the first 20 feature maps. Best viewed in color.}
\label{fig:5}
\end{figure*}
\subsection{Fine-Grained Classification}

The difficulty of fine-grained classification tasks is that even with different categories of objects, they are still very similar. This requires the network to have the ability to learn multiple and more accurate region features. In this section, we investigate the performance of SRP on fine-grained classification tasks.

We first analyze the results on CUB-200-2011 dataset, as shown in Table \ref{tab:4a}. Our method achieves strong performance with ResNet-50, which surpasses some deeper or larger network such as DenseNet-161~\cite{huang2017densely} or ResNet-110~\cite{he2016deep} more than 1\%. Also, compared with the method using extra annotation (FCAN~\cite{liu2016fully}), the method using multiple training stage (Part-RCNN~\cite{zhang2014part}), and the method using both extra annotation and multi-stage (RACNN~\cite{fu2017look}), our method outperforms them by 0.9\%, 4.0\% and 0.3\% respectively.

Our method also obtains the good performance on the Stanford Cars and Stanford Dogs, as shown in both Table \ref{tab:4b} and Table \ref{tab:4c}. On Stanford Cars, SRP outperforms all the comparison methods, except PA-CNN that uses extra annotation. On Stanfor Dogs, SRP surpasses the best result of other methods about 1.4\% in Table \ref{tab:4c}. Also, SRP outperforms its deeper or larger counterparts such as ResNet-110 and DenseNet-161 by about 1.0\% on Stanford Cars and 3.3\% on Stanford Dogs averagely. Furthermore, on these two datasets, SRP exceeds the efficient methods like DVAN~\cite{zhao2017diversified} and RAN~\cite{wang2017residual} by about 5.2\% and 2.5\% respectively, while DVAN and RAN both use extra annotation and multi-stage.


These facts indicate that SRP can obtain the significant improvement on fine-grained datasets, which even challenge the state-of-the-art results. It is worth mentioning that our method is a general method for the attention mechanisms. It may further improve the classification accuracy by combining with the better base network or some methods which are specifically for the fine-grained image classification.


\section{Discussion and Analysis}

\begin{figure}
\centering
  \includegraphics[width=8.3cm]{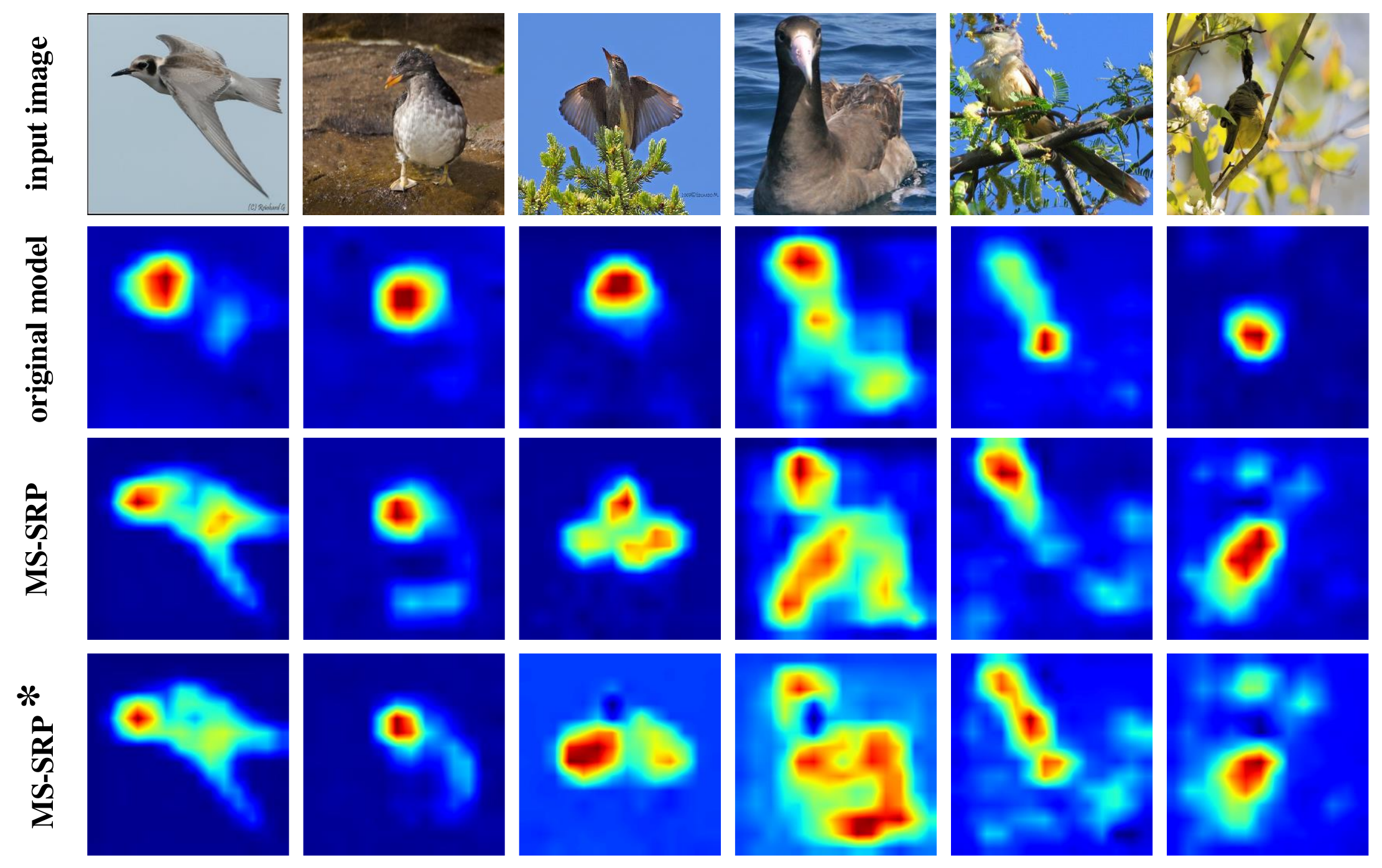}
  \caption{In \textbf{Fine-Grained} dataset, the Grad-CAM~\cite{selvaraju2017grad} visualization datasets for double branch block of attention ResNet50 model trained without SRP, trained with MS-SRP($M=5,\lambda=0.6$), and trained with MS-SRP$^*$($M=5,\lambda=0.2$). It indicates that SRP can promote the network to learn more detailed features of object, while SRP with too smal $\lambda$ will make the network pay attention to some unimportant fragment regions. Best viewed in color.}
  \label{fig:6}
\end{figure}


\noindent\textbf{Scheduled SRP.} The results in figure \ref{fig:2} shows that the scheduled SRP is superior to SRP with fixed $\lambda$ within a wide range of hyper-parameters. This indicates that the scheduled SRP is effective and practical. The possible reason is that the receptive field of shallow neurons in CNN are small. When SRP takes the fixed $\lambda$, the attention mechanism will receive a over-fragmented information about the feature map in shallow layer, which will in turn disturb the learning of the attention mechanism. Scheduled SRP avoided this problem by reducing $\lambda$ from 1 to the target value as the depth of layer increases.

\noindent\textbf{Feature maps analysis.} The purpose of SRP is to improve the representation of channel descriptors by increasing or widening the important responses in the feature map. Hence we output some feature maps of SRP-O-ResNet, SRP-D-ResNet and D-ResNet to analyze the effects of SRP visually, as shown in Figure \ref{fig:5}. For SRP-O-ResNet, SRP only acts on the residual branch. For SRP-D-ResNet, SRP acts on both the residual and identity branchs. For D-ResNet, SRP is not used.
It can be seen that in the residual branch, the feature map of SRP-O-ResNet and SRP-D-ResNet contains more and wider responses than D-ResNet (double branch attention of ResNet-50). In the identity branch, the feature map of SRP-D-ResNet contains more and wider responses than both SRP-O-ResNet and D-ResNet. These facts indicate that due to the effect of SRP, the feature maps from corresponding branch will have more and wider object responses.

\noindent\textbf{SRP on fine-grained Recognition.} In order to investigate the reason why SRP is more effective in fine-grained classification tasks, we use Grad-CAM to compare the visualization results of network train with or without SRP in CUB-200-2011. As we can see in Figure \ref{fig:6}, SRP can promote the network to focus on more details of the bird, such as the tip of the wing, the tail or the claw, while the network without SRP mostly tends to focus on one region. This indicates that SRP can promote the network to learn more detailed features of object, which becomes the key to the success of SRP in fine-grained classification tasks. It can be also observed that the network will pay attention to some unimportant fragment regions when $\lambda$ of SRP is too small. We conjecture that a too small $\lambda$ will cause SRP to obtain the over-fragmented information about the feature map, which may affect the performance of the attention mechanism.

\section{Conclusion}

In this paper, we propose a new method called Stochastic Region Pooling(SRP) for channel-wise attention networks. SRP stochastically selects the region from the feature map to extract descriptor in the training stage, promoting convolutional layer have more important feature responses, and making the network to focus on more and wider spatially distributed regions. Besides, SRP is the general method that can be applied to attention network without modifying the network structure and increasing any additional parameters. Our experiments show that the channel-wise attention network trained with SRP can achieve significant performance improvements on various image classification tasks and challenge the state-of-the-art methods. It is also proved that gradually decreasing the scale ratio of region as the depth of layer leads to the better accuracy. 



{\small
\bibliographystyle{ieee}
\bibliography{bibliography}
}

\clearpage
\section*{\Huge{Appendix}}
%

\subsection*{1. Area ratio of region in SRP}
\label{A1}
\begin{figure}[!htp]
\centering 
  \includegraphics[width=8cm]{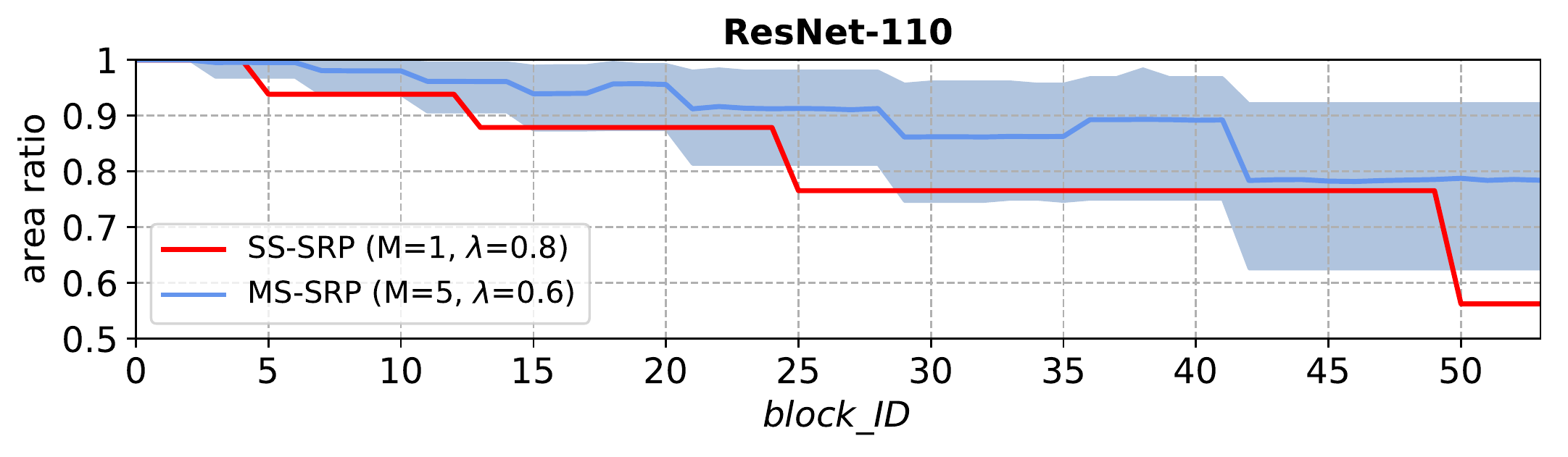}
  \includegraphics[width=8cm]{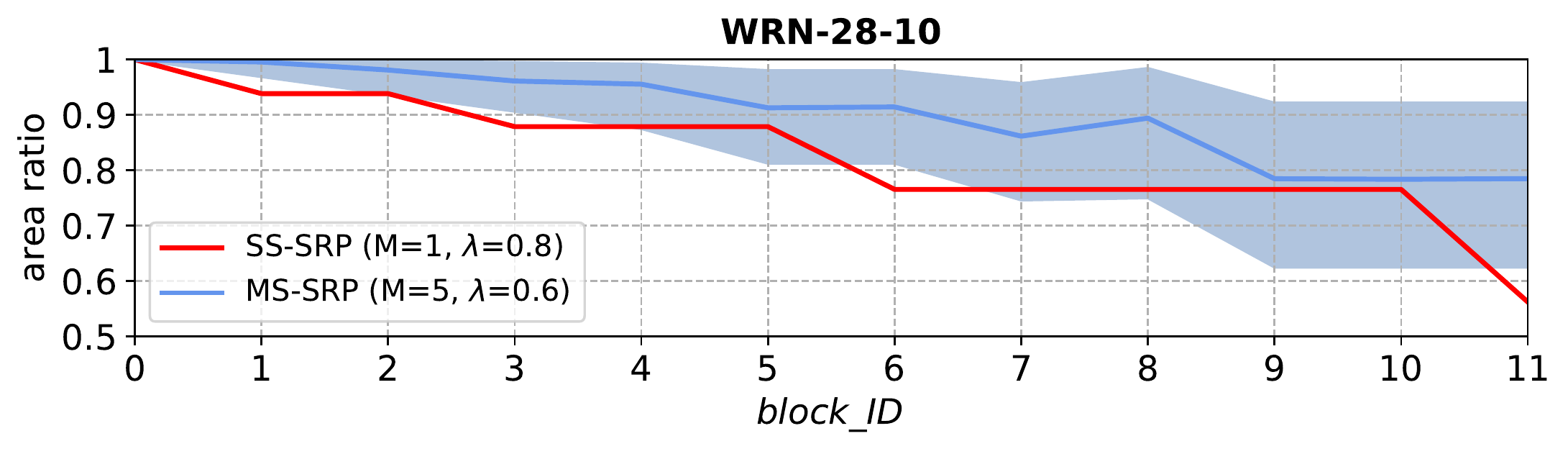}
  \includegraphics[width=8cm]{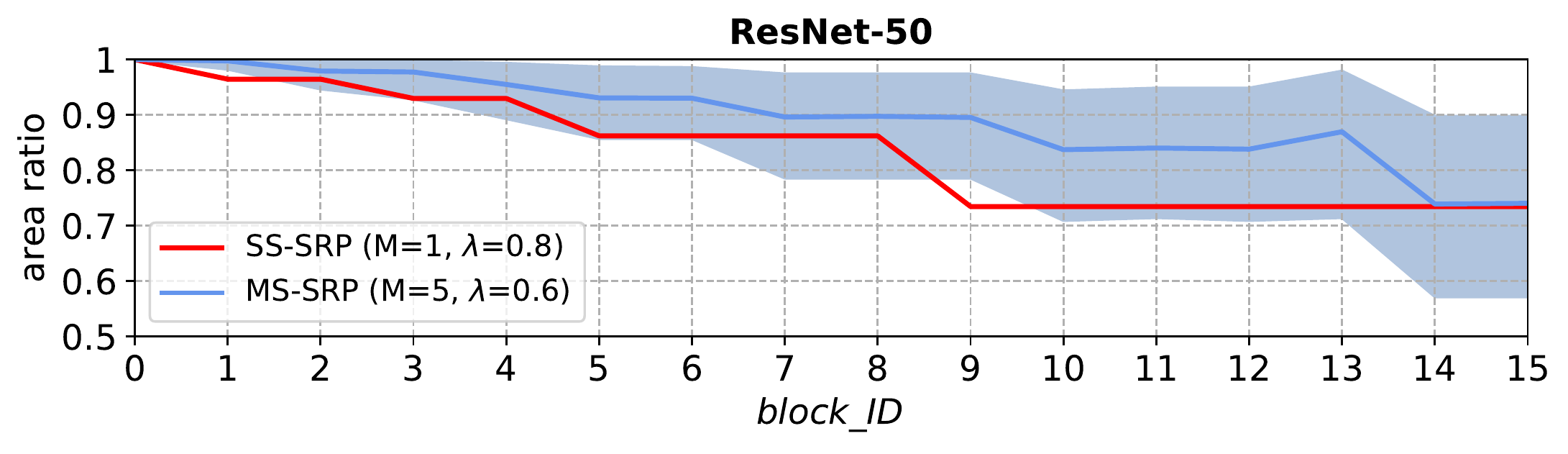}
\caption{The curve of region area ratio on different depth of netwrok blocks, where the region is seleted by SRP. The {\color{red}red line} is the value of the region area ratio in SS-SRP. The {\color{blue}blue line} is the mean of the region area ratio in the MS-SRP, and the value of area ratio in MS-SRP has a probability of 95\% on the {\color{blue}blue shadow}. Best viewed in color.}
\label{fig:7}
\end{figure}

The area ratio of region selected by SRP is different under different hyper-parameters($M,\lambda$); the area ratio is equal to the area of region divided by the area of the feature map. Figure \ref{fig:7} shows the curve of the area ratio of region selected by SRP on different depths of network (ResNet-110, WRN-28-10 and ResNet-50). Under the sampe depth, the area ratio of region takes value in a large range in the MS-SRP, but is a fixed value in the SS-SRP. We speculate that this smoothing is one of the reasons that make MS-SRP better than SS-SRP and allows SRP steadily promote network learning. 
 
\subsection*{2. Feature maps of network block at different depths}
\label{A2}
\begin{figure}[!htp] 
\centering
  \includegraphics[width=7.5cm]{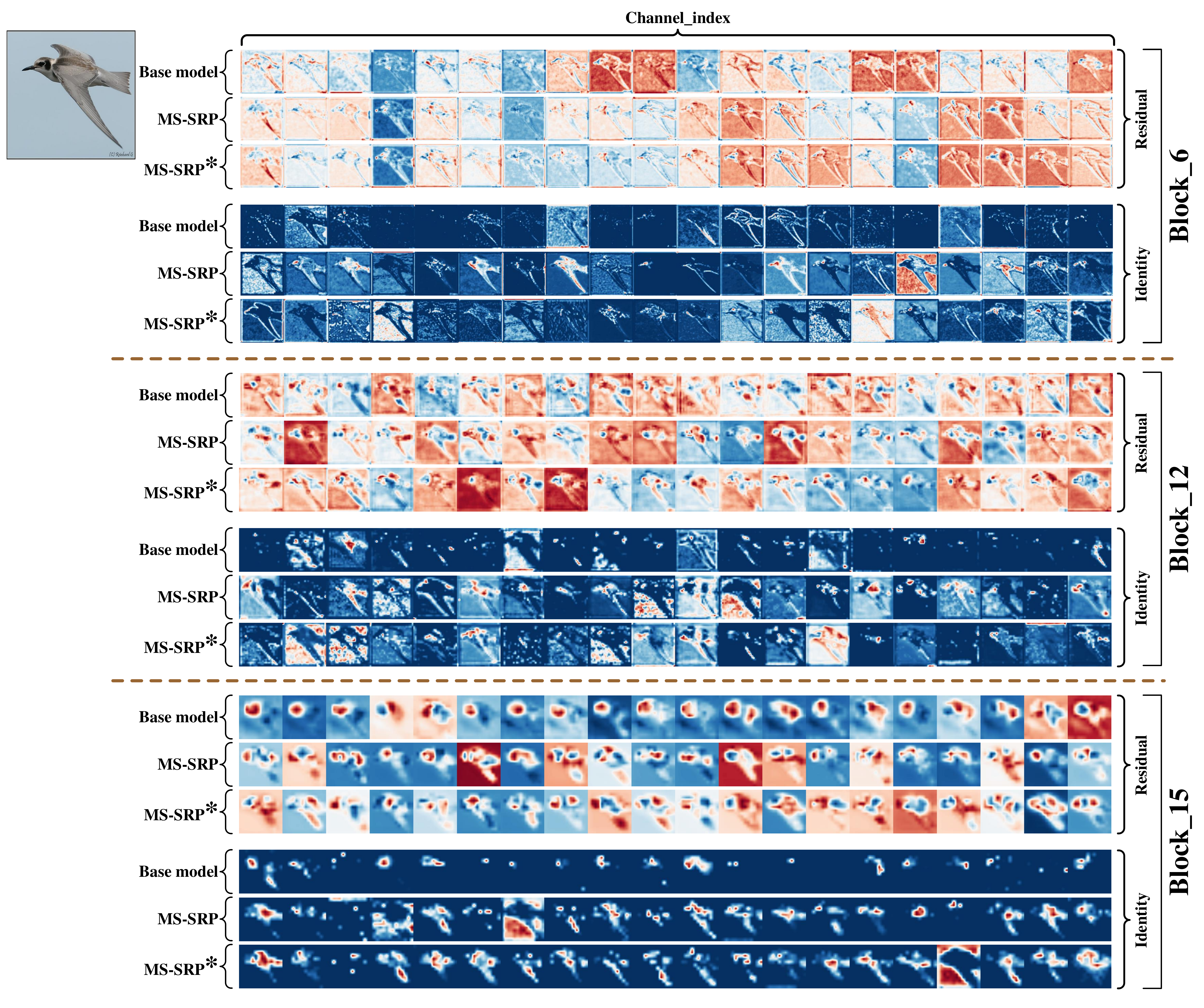}
  \includegraphics[width=7.5cm]{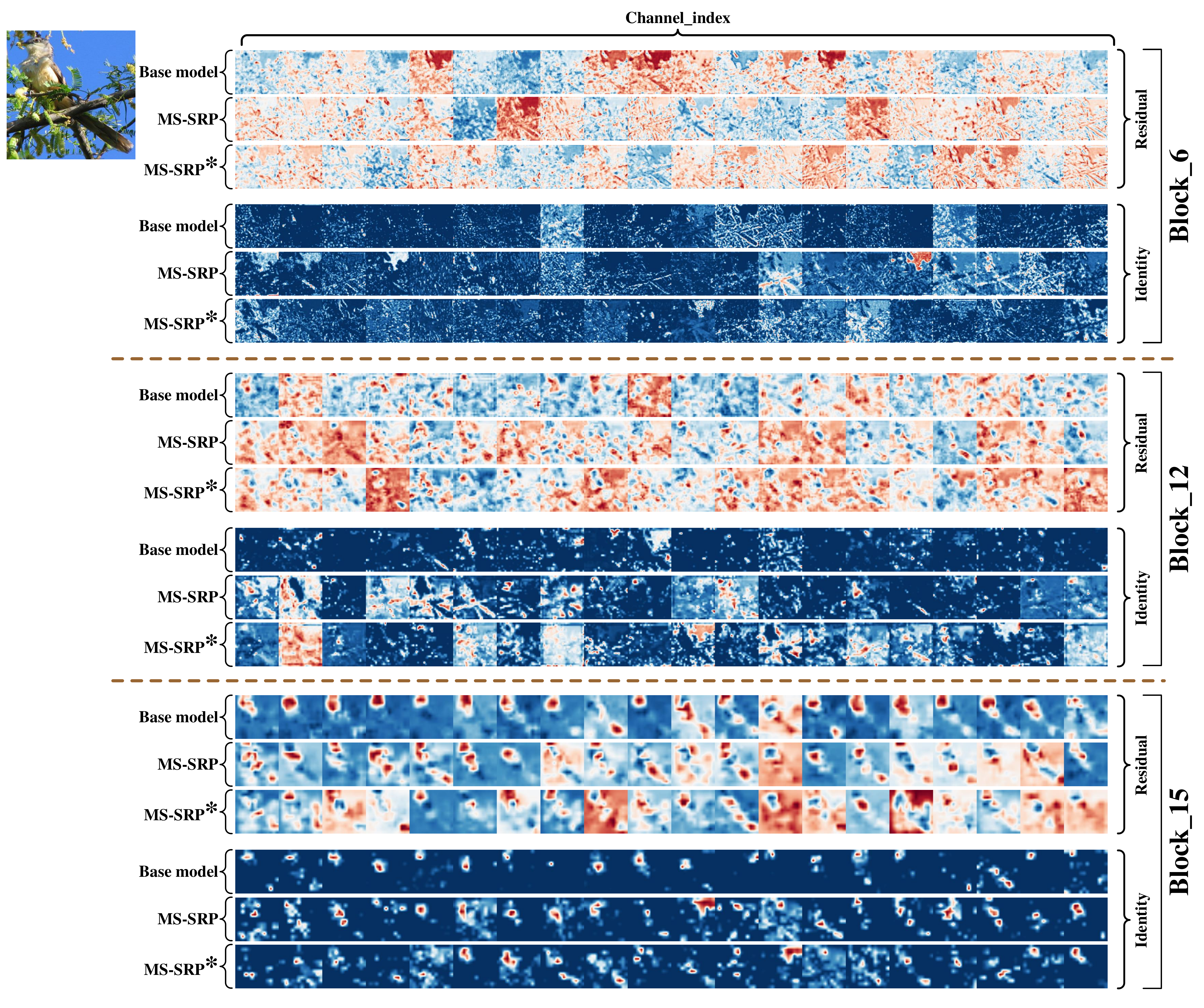}
  \includegraphics[width=7.5cm]{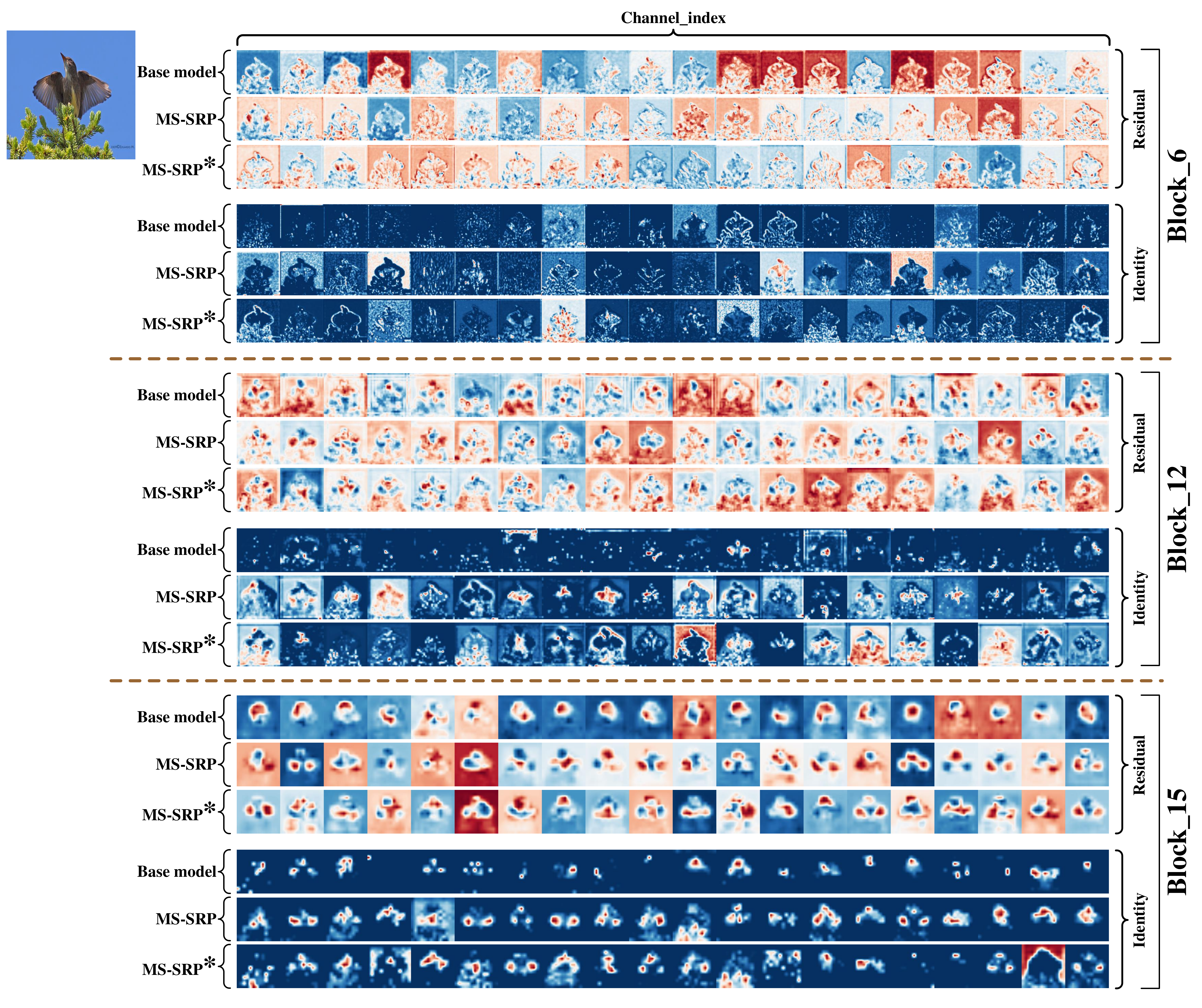}
  \caption{The visualization of feature maps from network block of different depths in \textbf{Fine-Grained} dataset. The Base model means D-ResNet-50, the MS-SRP means D-ResNet-50 trained with SRP($M=5, \lambda=0.6$) and the MS-SRP$^*$ means D-ResNet-50 trained with SRP($M=5, \lambda=0.2$). We only display the first 20 feature maps. Best viewed in color.}
\label{fig:8}
\end{figure}

Figure \ref{fig:8} shows the visualization of feature maps from the network which trained with or without SRP at different depth. We can observe that the deeper the network block, the more obvious the SRP characteristic that makes the feature map contains more and wider feature responses. At the same time, we found that too small $\lambda$ values tend to generate more but messy responses in deep layer.

\end{document}